\documentclass[pdflatex,sn-mathphys-num]{sn-jnl}

\usepackage{graphicx}%
\usepackage{multirow}%
\usepackage{amsmath,amssymb,amsfonts}%
\usepackage{amsthm}%
\usepackage{mathrsfs}%
\usepackage[title]{appendix}%
\usepackage{xcolor}%
\usepackage{textcomp}%
\usepackage{manyfoot}%
\usepackage{booktabs}%
\usepackage{algorithm}
\usepackage{algorithmic}
\usepackage{listings}%
\usepackage{adjustbox}
\usepackage{multicol}

\begin{document}

\title[Arabic Multi-Label Emotion Classification]{Improving Arabic Multi-Label Emotion Classification using Stacked Embeddings and Hybrid Loss Function}


\author*[1]{\fnm{Muhammad Azeem} \sur{Aslam}}\email{azeem@eurasia.edu}

\author[1]{\fnm{Wang} \sur{Jun}}\email{wangjun@eurasia.edu}

\author[2]{\fnm{Nisar} \sur{Ahmed}}\email{nisarahmedrana@yahoo.com}

\author[3]{\fnm{Muhammad Imran} \sur{Zaman}}\email{imranzaman.ml@gmail.com}

\author[1]{\fnm{Li} \sur{Yanan}}\email{lyn2541023277@qq.com}

\author[1]{\fnm{Hu} \sur{Hongfei}}\email{huhongfei@eurasia.edu }

\author[1]{\fnm{Wang} \sur{Shiyu}}\email{zai1245923417@outlook.com}

\author[1]{\fnm{Xin} \sur{Liu}}\email{liuxin@eurasia.edu}

\affil*[1]{\orgdiv{School of Information Engineering}, \orgname{Xi'an Eurasia University}, \orgaddress{\city{Xi'an}, \postcode{710071}, \state{Shaanxi}, \country{China}}}

\affil[2]{\orgdiv{Department of Computer Science (New Campus)}, \orgname{University of Engineering \& Technology}, \orgaddress{\city{Lahore}, \postcode{54000}, \state{Punjab}, \country{Pakistan}}}

\affil[2]{\orgdiv{Department of Computer Science}, \orgname{COMSATS University Islamabad (Lahore Campus)}, \orgaddress{\city{Lahore}, \postcode{54000}, \state{Punjab}, \country{Pakistan}}}


\abstract{In multi-label emotion classification, particularly for low-resource languages like Arabic, the challenges of class imbalance and label correlation hinder model performance, especially in accurately predicting minority emotions. To address these issues, this study proposes a novel approach that combines stacked embeddings, meta-learning, and a hybrid loss function to enhance multi-label emotion classification for the Arabic language. The study extracts contextual embeddings from three fine-tuned language models—ArabicBERT, MarBERT, and AraBERT—which are then stacked to form enriched embeddings. A meta-learner is trained on these stacked embeddings, and the resulting concatenated representations are provided as input to a Bi-LSTM model, followed by a fully connected neural network for multi-label classification.   To further improve performance, a hybrid loss function is introduced, incorporating class weighting, label correlation matrix, and contrastive learning, effectively addressing class imbalances and improving the handling of label correlations. Extensive experiments validate the proposed model's performance across key metrics and a staggering Jaccard accuracy of 0.81, F1-score of 0.67 and a hamming loss of 0.15. The class-wise performance analysis demonstrates the hybrid loss function's ability to significantly reduce disparities between majority and minority classes, resulting in a more balanced emotion classification. An ablation study highlights the contribution of each component, showing the superiority of the model compared to baseline approaches and other loss functions. This study not only advances multi-label emotion classification for Arabic but also presents a generalizable framework that can be adapted to other languages and domains, providing a significant step forward in addressing the challenges of low-resource emotion classification tasks.}




\maketitle

\section{Introduction}\label{sec1}
Social media platforms, particularly microblogging sites like Twitter, have revolutionized the way public sentiment and emotions are analyzed. Prior to social media, extracting valuable insights such as perceptions, opinions, and emotions from the public was a daunting task \cite{alqahtani2022emotion}. With millions of tweets posted daily, these platforms offer a rich source of data that reflects a wide range of human emotions and opinions \cite{wikarsa2015text,goldenberg2018emotional}. Understanding these emotional contents is crucial for providing context-related and personalized information, enhancing social movements, and improving various applications such as smart governance, education, and healthcare \cite{goldenberg2018emotional,milazzo2017exploiting}.\\
Emotion Classification (EC) leverages Natural Language Processing (NLP) techniques to extract subjective information from textual data \cite{wankhade2022survey,acheampong2020text}. This process is integral to multiple domains, including smart governance, business, and organizations, where it is applied to various data formats such as audio, text, and video. The theoretical foundation for EC is often grounded in psychological theories of emotion, such as Plutchik's "Wheel of Emotions" \cite{plutchik1980general} and Ekman's theory of basic emotions \cite{ekman1992argument}.\\
Automated EC, which includes Single-Label Emotion Classification (SLEC) and Multi-Label Emotion Classification (MLEC), has become essential due to the vast amount of text available online. MLEC is particularly advantageous as it identifies multiple emotions in a given text, addressing the limitations of SLEC which can only detect a single emotion per text.\\
In recent years, significant advancements in word embedding (WE) have contributed to the progress of MLEC. WE techniques represent words in dense, low-dimensional vectors that capture syntactic and semantic relationships, enabling better understanding and processing of human language by machines \cite{mikolov2013efficient,boudad2020exploring}. However, Arabic text presents unique challenges for WE due to its complex morphology and diverse dialects \cite{sabri2024improved}.\\
Addressing the issues of dataset imbalance and class discrimination in MLEC is critical. Contrastive Learning (CL) has shown promise in learning effective representations by maximizing the similarity between related data samples and distinguishing dissimilar ones \cite{gao2021simcse}. CL has achieved cutting-edge performance in capturing semantic representations of sentences \cite{zhou2022employing} and addressing challenges in distinguishing discrete and valence dimension-based emotions \cite{tursunov2019discriminating}.\\
Despite the progress in EC using various machine learning and deep learning approaches, Arabic MLEC faces challenges due to the small and imbalanced datasets available. Traditional loss functions like Binary Cross Entropy (BCE) are inadequate for handling class imbalance effectively \cite{bengio2013representation}. To mitigate this, Focal Loss (FL) was introduced to focus on hard-to-classify samples \cite{lin2023effective}. Recent studies have proposed Multi-label Contrastive Focal Loss (MCFL), which combines FL with CL to enhance performance in imbalanced datasets \cite{zheng2021multi,chen2022personalized}.\\
Deep learning models such as Bidirectional Long Short-Term Memory (Bi-LSTM) and Bidirectional Gated Recurrent Unit (Bi-GRU) have proven effective for NLP tasks. However, ensemble techniques, particularly stacking ensembles, offer improved predictions and generalization by combining outputs from multiple models \cite{haralabopoulos2020ensemble}. Stacking ensembles utilize diverse models to enhance accuracy and robustness, making them suitable for complex tasks like Arabic MLEC.\\
This research proposes an MLEC model by employing stacked ensemble word embeddings derived from pre-trained language models including MARBERT, AraBERT, and ArabicBERT. Our objective is to harness the power of these models to generate high-contextual embeddings tailored for Arabic texts. Furthermore, we address the inherent challenges of dataset imbalance and class discrimination through the utilization of a hybrid loss function. This Arabic MLEC model aims to advance the field of Arabic MLEC by providing a robust and reliable model capable of enhancing emotion detection in Arabic tweets.\\

The main contributions of this study are outlined as follows:
\begin{enumerate}
    \item Presented stacking approach for contextualized word embeddings by fine-tuning Arabic language models (MARBERT, AraBERT, and ArabicBERT) to facilitate high contextual embedding generation for Arabic MLEC.
    \item Introduced a hybrid loss function by combining class weights, label correlation and contrastive learning, tailored for multi-label text classification, addressing challenges related to dataset imbalance and class discrimination.
    \item Trained a second stage classifier as meta-learner which perform sequence learning using Bi-LSTM followed by multi-label classification learning using fully-connected network.
\end{enumerate}
\section{Related Work}
Emotion classification (EC) is a pivotal area of research in natural language processing (NLP) with the goal of distinguishing between different emotional or psychological states expressed in text \cite{nandwani2021review}. While substantial progress has been made in EC for high-resource languages like English, research focused on Arabic remains limited.
\subsection{Multi-Label Emotion Classification for Arabic Language}
Arabic MLEC has emerged as a significant research area attracting attention from various scholars aiming to address the unique challenges posed by Arabic text data. This section provides an overview of recent advancements in Arabic MLEC methodologies, with a focus on studies leveraging the SemEval-2018 task 1-Ec-Ar dataset \cite{mohammad2018semeval} as a benchmark. Several studies have contributed to the advancement of Arabic MLEC, each employing diverse methodologies to tackle the complexities inherent in Arabic text emotion analysis. Notably, the SemEval-2018 dataset has served as a common benchmark, facilitating comparative analyses and informing the development of novel approaches. \\
Lin et al. \cite{lin2023effective} introduced contrastive learning techniques to enhance MLEC performance, achieving competitive F-macro and F-micro scores. Mansy et al. \cite{mansy2022ensemble} explored the effectiveness of MARBERT, BiLSTM, and Bi-GRU models, demonstrating notable accuracy and precision in emotion detection tasks. Khalil et al. \cite{khalil2021deep} leveraged Bi-LSTM AraVec/CBOW models, showcasing competitive performance in terms of Fmicro, precision, recall, and Jaccard accuracy.\\
Elfaik et al. \cite{elfaik2021combining} proposed a hybrid approach combining AraBERT word embeddings with attention-based LSTM and BiLSTM models, achieving high accuracy in MLEC tasks. Abdelali et al. \cite{abdelali2023benchmarking} introduced the QARiB model and a hybrid feature model, yielding respectable F-macro scores in emotion classification. Additionally, Menai et al. \cite{alswaidan2020hybrid} presented a hybrid feature-based model and deep feature-based model, demonstrating strong performance in terms of F-micro and Jaccard accuracy. Samy et al. \cite{samy2018context} employed a context-aware GRU model, achieving competitive scores in emotion classification tasks.\\
The diverse range of methodologies and performance metrics employed in these studies \cite{lin2023effective,mansy2022ensemble,khalil2021deep,elfaik2021combining,abdelali2023benchmarking,alswaidan2020hybrid,samy2018context} underscore the complexity of Arabic MLEC and the need for innovative approaches to address it. Furthermore, the utilization of the SemEval-2018 dataset as a benchmark has facilitated comparative analyses and enabled researchers to build upon existing methodologies.\\
\subsubsection{Arabic Emotion Datasets}
In addition to methodological advancements, the development of Arabic emotion datasets has played a crucial role in facilitating research in this domain. Table \ref{tab:dataset} provides a summary of existing Arabic emotion datasets, highlighting key attributes such as size, annotation type, accessibility, and emotion coverage. Despite the availability of these datasets, challenges related to accessibility and balance in emotion representation persist, underscoring the need for further efforts to address these limitations.
\begin{sidewaystable}
\caption{Summary of Arabic Emotions Datasets}
\label{tab:dataset}
\begin{tabular*}{\textheight}{@{\extracolsep\fill}llllp{210pt}}
\toprule
\multicolumn{1}{|c|}{\textbf{Authors/Year}} & \multicolumn{1}{c|}{\textbf{Dataset}} & \multicolumn{1}{c|}{\textbf{Size}} & \multicolumn{1}{c|}{\textbf{Type/Task}} & \multicolumn{1}{c|}{\textbf{Emotions}} \\ 
\midrule
Althobaiti, (2023)                          & ArPanEmo                              & 11,128                             & DA/SLE                                  & Disgust, Happiness, Sadness, Confusion, Surprise, Anticipation, Optimism, Pessimism, Neutral          \\ 
Shakil et al. (2021)                        & AEELex                                & 35,383                             & DA/SLE                                  & Anger, anticipation, disgust, fear, joy, sadness, surprise, trust                                     \\ 
Azad et al. (2021)                          & ExaACE                                & 20,050                             & DA/MLE                                  & Anger, Anticipation, Love, Fear, Disgust, Joy, Sadness, Acceptance, Surprise, Neutral                 \\ 
Teahan (2019)                               & IAEDS                                 & 1,366                              & DA/SLE                                  & Anger, Disgust, Happiness, Fear, Surprise, Sadness                                                    \\ 
Mohammad et al. (2018)                      & SemEval 2018                          & 4,381                              & DA/MLE                                  & Anger, Anticipation, Disgust, Fear, Joy, Love, Optimism, Pessimism, Sadness, Surprise, Trust, Neutral \\ 
ElBeltagy et al. (2018)                     & ETDA                                  & 10,065                             & DA/SLE                                  & Joy, Sadness, Anger, Love, Fear, Sympathy, Surprise, None                                             \\ 
Mohammad et al. (2018)                      & SemEval 2018                          & 5,600                              & DA/SLE                                  & Joy, Anger, Sadness, Fear                                                                             \\ 
Alhuzali et al. (2018)                      & LAMA-DIST                             & 7,268                              & DA/SLE                                  & Anger, Anticipation, disgust, Fear, Joy, Sadness, Surprise, Trust                                     \\ 
Abdul-Mageed et al. (2016)                  & DINA                                  & 3,000                              & DA/SLE                                  & Anger, Disgust, Fear, Happiness, Sadness, Surprise                                                    \\ \botrule
\end{tabular*}
\end{sidewaystable}
Therefore, the advancements in Arabic MLEC methodologies and the availability of benchmark datasets have significantly contributed to the progress of research in this field. However, continued efforts are required to overcome existing challenges and further enhance the effectiveness of Arabic MLEC systems.
\subsection{Contextual Embeddings with Pre-Trained Language Models}
Language processing is a complex task, requiring computational techniques to convert words into numerical formats that machines can understand. This process, known as Word Embedding (WE) or Word Representation, is critical in NLP activities, allowing texts to be represented in vector spaces \cite{asudani2023impact}. WEs play a vital role in various NLP tasks, including machine translation, part-of-speech tagging, sequence labeling, named entity recognition, speech processing, and text classification \cite{mao2019sentiment,fang2022knowledge}. These embeddings capture the semantic meaning of words within text sequences, assigning numerical representations to words with similar meanings.\\
The features employed in classification tasks have a significant impact on how well supervised methods perform in EC \cite{mao2019sentiment}. Frequency-based methods like N-grams, Bag of Words, and Term Frequency-Inverse Document Frequency are examples of traditional WE techniques whereas GloVe \cite{pennington2014glove}, word2vec \cite{mikolov2013distributed} and FastText are the neural networks based WE techniques \cite{joulin2016fasttext}. These earlier models, however, struggled to capture complex semantic dependencies and contextual information, often representing words with single, context-independent vectors \cite{li2019dice,al2021evolution}.\\
Context-aware Pretrained Language Models (PLMs), like Bidirectional Encoder Representations from Transformers (BERT) \cite{devlin2018bert}, have been developed as a result of recent developments in NLP. These models, utilizing transfer learning, have significantly improved performance across various NLP tasks. Unlike earlier WE models, PLMs like BERT excel in capturing complex semantic dependencies and contextual information by using transformer architecture for sub-word learning \cite{devlin2018bert}.\\
Numerous studies have compared the efficiency of different WE techniques across various domains. Al-Twairesh et al. \cite{al2021evolution} highlighted that while static models have shown progress, they suffer from contextual information deficiencies and out-of-vocabulary (OOV) issues. David et al. \cite{david2021comparison} compared static WE techniques using RNN and CNN for text classification, finding that FastText outperformed GloVe. Moreover, reducing batch size improved accuracy, demonstrating the significance of training parameters in model performance.\\
The stacking of WE techniques involves combining multiple embeddings to leverage their strengths. According to Muromägi et al. \cite{muromagi2017linear}, stacked WE techniques can cancel out random noise from individual models and reinforce useful patterns. This approach is particularly beneficial in languages with limited textual data. Stacked WE techniques have shown improved performance over single models across various NLP tasks \cite{speer2016ensemble}.\\
In the context of Arabic NLP, several BERT models such as AraBERT \cite{antoun2020arabert}, ArabicBERT-Large \cite{safaya2020kuisail}, and MARBERT \cite{abdul2020arbert} have been developed. These models, trained on large-scale Arabic corpora, have demonstrated impressive performance in tasks like sentiment analysis and text classification \cite{aoumeur2023improving}. AraBERT has shown superior performance in Arabic language understanding, while MARBERT, which focuses on dialectal Arabic, captures diverse Arabic dialects effectively \cite{fadel2019pretrained}.\\
Systematic reviews and comparative studies have underscored the effectiveness of BERT models in Arabic text classification. Alammary et al. \cite{alammary2022bert} reviewed 66 studies, highlighting the efficiency of these models in various NLP tasks. Berrimi et al. \cite{berrimi2023attention} compared existing Arabic text classification approaches, supporting the use of BERT models for improved performance \cite{elbedwehy2023improved}.\\
Despite the notable performance of individual BERT models, the stacking of these models has shown further improvements. Studies such as those by Kamr et al. \cite{kamr2022akabert} and Chouikhi et al. \cite{chouikhi2021stacking} have demonstrated that ensemble BERT models outperform single BERT techniques. However, no research has yet applied stacking techniques combining AraBERT, MARBERT, and ArabicBERT for feature extraction in Arabic Multilingual Emotion Classification (MLEC), despite the promising potential indicated by individual model performances.\\
The literature indicates that contextual embeddings using PLMs like BERT have revolutionized NLP by providing more accurate and context-aware word representations. The stacking of WE techniques further enhances model performance, particularly in languages with limited textual data.
\subsection{Loss Functions for Imbalanced Data}
Dataset imbalance, particularly prevalent in MLEC, presents significant challenges in achieving balanced and accurate model performance. According to Padma et al. \cite{padma2014classification} and Kumar et al. \cite{kumar2021classification}, a dataset is considered balanced when it contains an equal number of data points for all classes; otherwise, it is deemed imbalanced. This imbalance is especially noticeable in publicly available datasets. To address this, researchers \cite{durand2019learning,dong2020focal,jiang2021text} have proposed various loss functions designed to mitigate the effects of class imbalance, such as the Multi-label Focal Loss  which addresses class imbalance by focusing more on hard-to-classify, minority class examples and reducing the emphasis on well-classified, majority class examples. In multi-label settings, it applies this weighting independently to each label, improving performance across both majority and minority classes.\\
Binary Cross Entropy (BCE) loss is a commonly used loss function for multi-label text classification tasks, including MLEC. However, BCE is highly susceptible to label imbalance, often influenced by the prevalence of dominant classes or negative instances \cite{durand2019learning}. To address this, Focal Loss (FL) was proposed by Lin et al. \cite{lin2017focal} to reshape the standard cross-entropy loss, placing more emphasis on hard and misclassified samples. Dong \cite{dong2020focal} highlights FL's efficacy in focusing on difficult instances more effectively than classic cross-entropy loss. Jiang et al. \cite{jiang2021text} assert that while FL was initially applied in computer vision, it also shows reputable performance in text mining. Consequently, due to their success in other domains, loss functions like FL have been adapted for use in NLP.\\
Researchers such as Li et al. \cite{li2019dice} and Sun et al. \cite{sun2019drug} have applied these adapted loss functions to NLP tasks. For instance, Sun et al. \cite{sun2019drug} applied FL to an information extraction model analyzing biomedical literature, reporting effective results in addressing imbalance. Similarly, Zhu et al. \cite{zhu2020speech} utilized FL in speech emotion recognition, improving performance on the Interactive Emotional Dyadic Motion Capture (IEMOCAP) dataset.\\
While FL has demonstrated success in various tasks, its effectiveness is more pronounced in single-label classification. According to Ridnik et al. \cite{ridnik2021asymmetric}, treating positive and negative samples equally in multi-label classification, as FL does, can accumulate greater loss gradients from negative samples, reducing the importance of rare positive examples. This limitation has led to the development of Multi-label Focal Loss (MFL) to better accommodate multi-label classification tasks, including MLEC.\\
Deng et al. \cite{deng2020multi} applied MFL to address class imbalance in MLEC on the RenCECps and NLPCC2018 datasets, achieving better performance than state-of-the-art methods. Similarly, Bendjoudi et al. \cite{bendjoudi2020audio} used MFL in their multi-label emotion recognition model, achieving remarkable results and outperforming state-of-the-art methods for minority classes.\\
Contrastive learning (CL) has also been explored as a method for handling data imbalance. CL learns generalized representations by capturing similarities within a class and contrasting them with other classes \cite{khosla2020supervised}. This approach addresses the fine-grained feature spaces of similar emotions, enhancing generalization ability \cite{hu2023multi}. Researchers have proposed various CL methods to improve emotion discrimination, and these methods have shown promise in handling imbalanced data.\\
Combining Focal Loss and Contrastive Learning has led to the development of Multi-label Contrastive Focal Loss (MCFL). This combination leverages the strengths of both methods: FL's focus on hard instances and CL's ability to differentiate between similar and dissimilar data points. Studies by Chen et al. \cite{chen2022personalized} and Zheng et al. \cite{zheng2021multi} in computer vision have shown that MCFL outperforms state-of-the-art methods in imbalanced datasets. Lin et al. \cite{lin2023effective} further explored the efficiency of CL in multi-label text classification, including Arabic MLEC, using BERT-base-Arabic for fine-tuning. Their study proposed five novel multi-label contrastive losses, with the Stepwise Label Contrastive Loss (SLCL) achieving the highest FMacro and Strict Contrastive Loss (SCL) achieving the highest FMicro and Jaccard Score. These studies collectively highlight the effectiveness of combining loss functions to minimize the effects of class imbalance, allowing the model to efficiently learn multi-label classification. This research gap presents an opportunity to formulate a joint loss function that combines contrastive learning with variants addressing class imbalance, thus justifying the need for a hybrid loss function.
\section{Methodology}\label{sec:methodology}
\begin{figure}[h!]
    \centering
    \includegraphics[width=\textwidth]{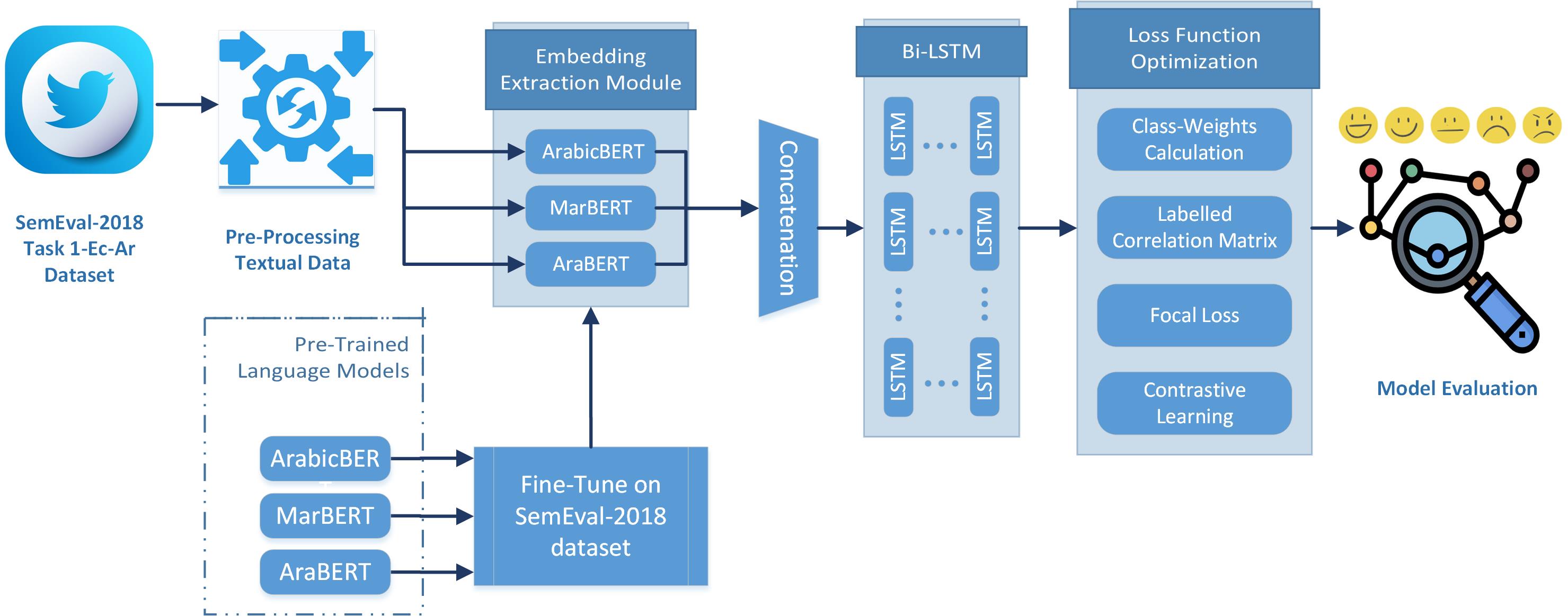}
    \caption{Framework of the proposed approach depicting various stages}
    \label{fig:framework}
\end{figure}

This study adopts a systematic paradigm which views computer science programs as entities analogous to mental processes, integrating both formal deduction and scientific experimentation to acquire a priori and posteriori knowledge \cite{eden2007three}. This approach is particularly prevalent in the field of artificial intelligence, ensuring rigorous and systematic investigation.
The research focuses on the implementation of Arabic MLEC. The framework designed to achieve the research objectives is depicted in Figure \ref{fig:framework}. This framework outlines the procedural steps and methodological tools used, including data pre-processing, model development, and evaluation metrics. Guided by the scientific paradigm, the approach systematically address the challenges associated with Arabic MLEC, validating the findings through both theoretical and empirical means. This alignment with established practices in AI research ensures the robustness and credibility of the study's results.
\subsection{Dataset Description} \label{dataset_description}
\begin{figure}[h!]
    \centering
    \includegraphics[width=0.9\textwidth]{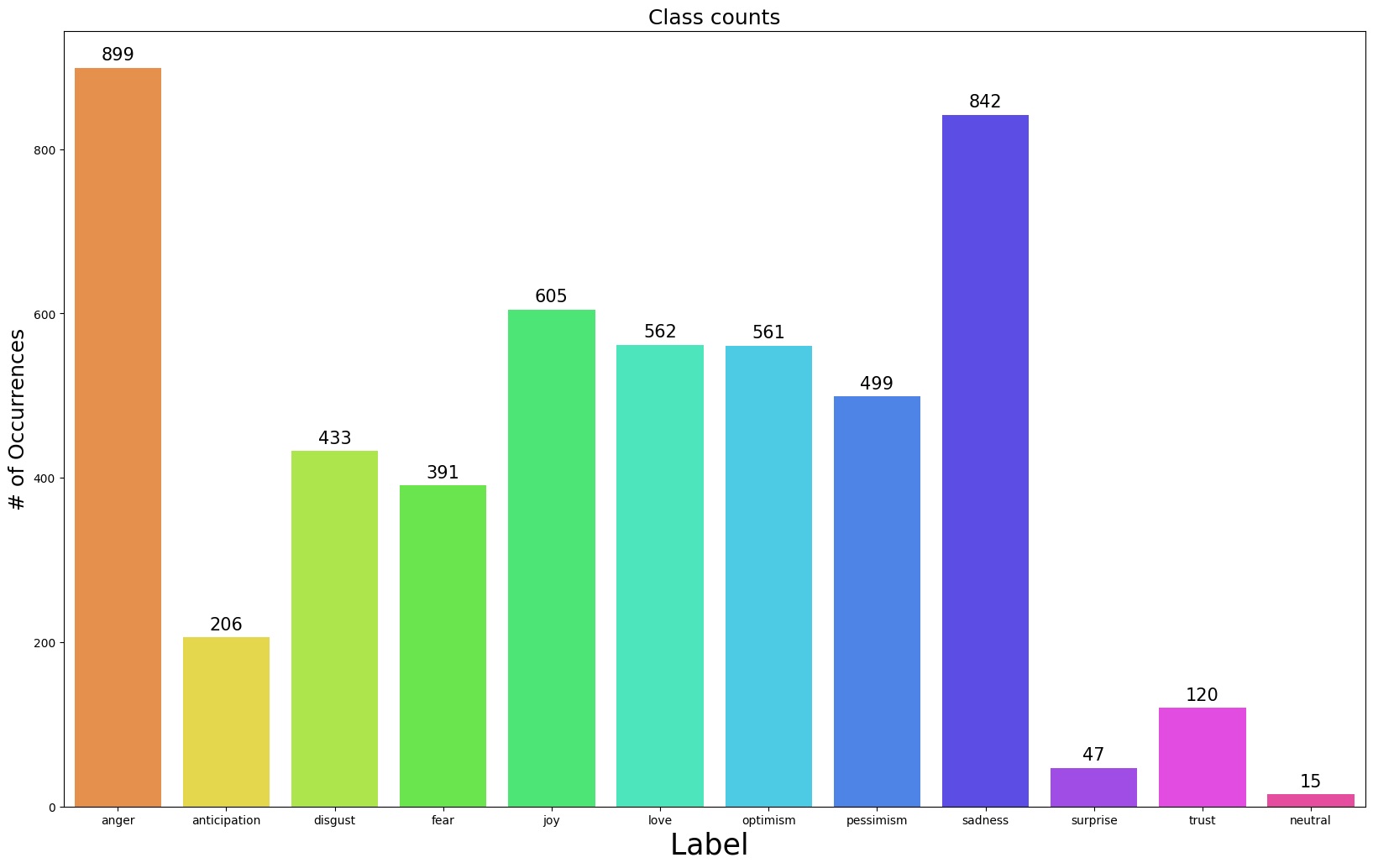}
    \caption{Distribution of instances in each emotion category}
    \label{fig:emotion_distribution}
\end{figure}

Utilizing existing datasets for text based emotion classification is common among researchers. However, many online datasets suffer from significant class imbalances, impeding the performance of supervised learning models \cite{alrasheedy2022text}. Such imbalances often lead models to favor majority classes, resulting in suboptimal predictions for minority classes.\\
In response to this issue, this study aims to address class imbalance in existing datasets, with a particular focus on the SemEval-2018 Task 1-Ec-Ar dataset curated by Mohammad et al. \cite{mohammad2018semeval}. This dataset has been extensively employed in prior research on MLEC and provides Arabic text samples with annotated emotion labels, making it a valuable resource for developing and testing advanced MLEC models.\\
The dataset comprises eleven emotion classes, including anger, fear, anticipation, joy, disgust,love, pessimism, optimism, trust, surprise and sadness \cite{mohammad2018semeval}. The dataset was collected using the Twitter API, with search queries aligned to tweets containing emotional words. Annotation reliability was determined using Best Worst Scaling (BWS), with each tweet labeled with '1' indicating the presence of the corresponding emotion class and '0' indicating its absence. Table \ref{tab:dataset_snippet} provides some sample instances containing Tweets and the multi-label emotion class identified for each Tweet.  \\
\begin{table}[h!]
    \centering
    \caption{Tweets along with their multi-label emotion classes}
    \begin{tabular}{c}
        \includegraphics[width=\textwidth]{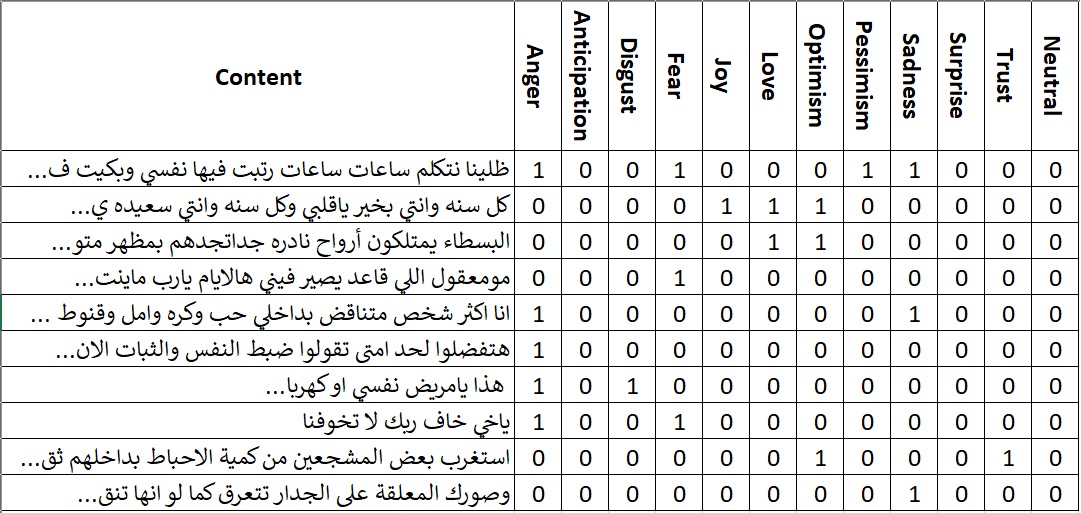} \\
    \end{tabular}
    \label{tab:dataset_snippet}
\end{table}
Despite its usefulness, the SemEval-2018 Task 1-Ec-Ar dataset suffers from imbalanced emotion label distributions, a common challenge in real-world datasets. Figure \ref{fig:emotion_distribution} provides distribution of instances in each emotion class indicating class imbalance. This imbalance presents a significant obstacle for ML and DL models, particularly in accurately predicting underrepresented emotion classes.
This study proposes use of an optimized loss function created especially to address the problem of class imbalance in order to mitigate this challenge. In order to improve MLEC model performance on the SemEval-2018 Task 1-Ec-Ar dataset and enable more accurate and balanced emotion classification, the research will concentrate on this aspect. This approach improves the accuracy and aplicability of emotion classification in Arabic text while also helping to build stronger ML and DL models.
\subsection{Pre-Processing}
\begin{figure}[h!]
    \centering
    \includegraphics[width=\textwidth]{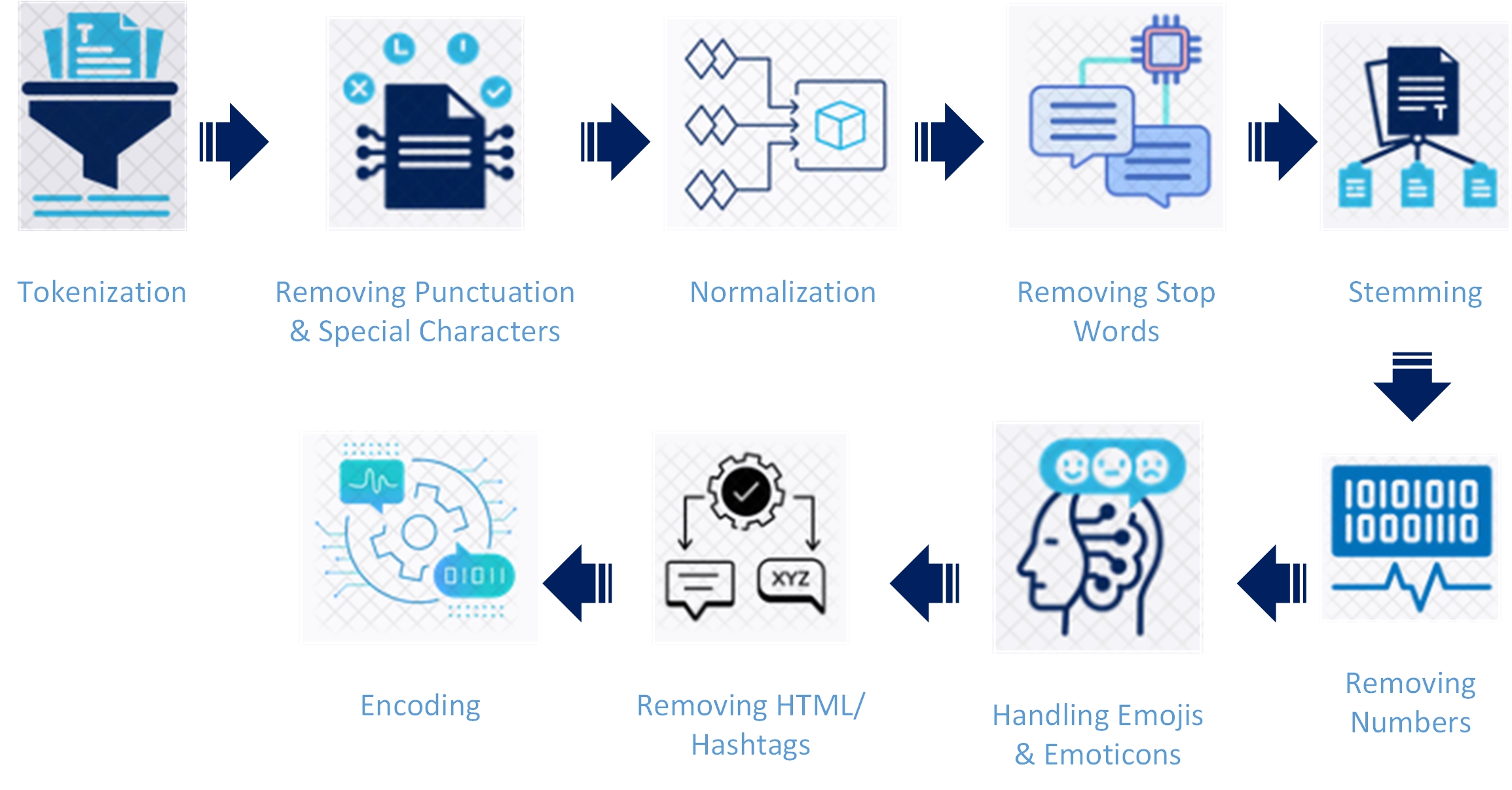}
    \caption{Framework of the proposed approach depicting various stages}
    \label{fig:preprocessing}
\end{figure}

Text data preprocessing is a critical step in creating robust machine learning models for NLP tasks like MLEC. The preprocessing involved several steps to handle various text formatting issues, particularly focusing on the treatment of punctuation, emojis, emoticons, and normalization of Arabic text. Figure \ref{fig:preprocessing} provides a flowchart depicting the steps and their sequence to perform preprocessing of the text.\\
First, both English and Arabic punctuation marks were replaced with whitespace. This ensures that punctuation does not interfere with the tokenization process. Replacing punctuation helps in maintaining the integrity of the words and prevents any misinterpretation during the model training phase.
Next, emojis and common emoticons were replaced with their corresponding textual tokens based on commonly available emoji and emoticon dictionaries. This approach preserves the emotional context conveyed by these symbols, which are prevalent in informal text. By converting emojis and emoticons to textual tokens, the preprocessing retains valuable emotional cues that might otherwise be lost if left untreated or removed from the data. This step ensures that the sentiment and emotional nuances embedded in the original text are captured and utilized effectively during model training.\\
The text was then stripped of English letters and numbers, focusing solely on the Arabic content. This step is essential for eliminating any extraneous characters that could skew the analysis towards non-Arabic language patterns. Similarly, Arabic numbers were also replaced with whitespace to ensure that the numerical data does not affect the linguistic model.\\
Normalization of Arabic text was another critical step. This process involved standardizing different forms of the same letter and removing diacritics. Normalization ensures that variations in spelling or character forms do not lead to misclassification or redundancy in the dataset. It helps in treating different forms of the same word as a single entity, thereby enhancing the consistency of the text data.\\
Specific characters such as dashes and decorative symbols were replaced with whitespace. These characters can often appear in social media texts and might not carry significant meaning related to the content. Removing them helps in focusing on the actual textual information. Similarly, backslashes were replaced with whitespace to prevent any potential issues in text parsing. Repeated characters were reduced to a single occurrence. This is particularly important in languages like Arabic, where character repetition can occur frequently for emphasis. Reducing repeated characters helps in normalizing the text and reducing redundancy. Extra whitespaces were removed to ensure that the text data is clean and compact. Additionally, standalone single characters were eliminated as they often do not contribute significant meaning and can be considered noise in the text data.\\
Further, specific patterns such as words starting with \includegraphics[height=\fontcharht\font`\B]{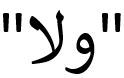} were addressed. The replacement of such patterns ensures that common prefixes are treated uniformly. This helps in maintaining the grammatical and contextual integrity of the sentences.\\
Finally, after these preprocessing steps, the text data was ready for extraction of word embeddings using pre-trained models from Hugging Face, specifically AraBERT, ArabicBERT and MARBERT. These models leverage state-of-the-art language representations and are well-suited for processing Arabic text, ensuring that the input data for the machine learning and deep learning models is of high quality and consistency. This comprehensive preprocessing pipeline is crucial for training effective models for Arabic MLEC, enhancing the reliability and accuracy of emotion detection in Arabic text.
\subsection{Stacked Embeddings Extraction}
\begin{figure}[h!]
    \centering
    \includegraphics[width=\textwidth]{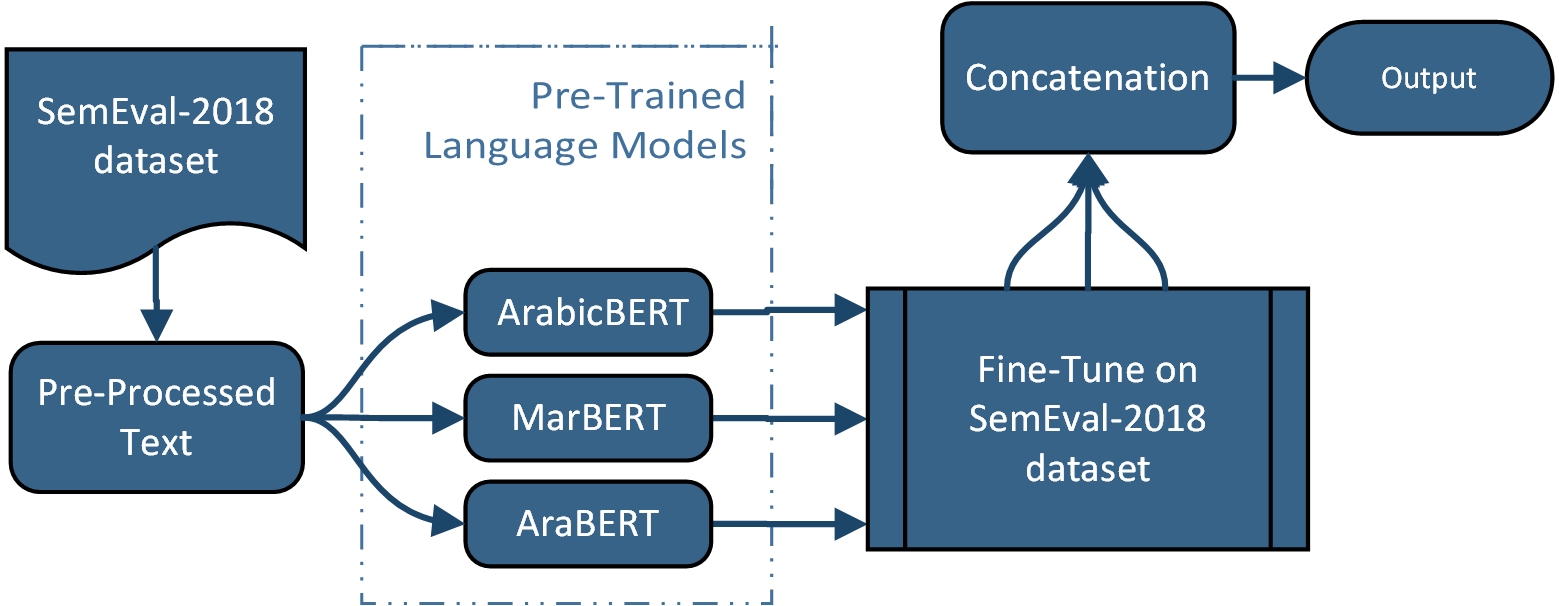}
    \caption{Contextualized embeddings generation for stacking ensemble}
    \label{fig:embeddings}
\end{figure}

To generate highly contextualized embeddings from Arabic text for multi-label emotion classification (MLEC), a stacked embedding extraction approach is introduced. This method leverages three pre-trained language models specifically tailored for the Arabic language: ArabicBERT, MarBERT, and AraBERT. Figure \ref{fig:embeddings} illustrates the process of extracting embeddings from the preprocessed text.
\subsubsection{Fine-Tuning the Language Models}
The fine-tuning process is performed using the pre-processed SemEval-2018 dataset, which contains Arabic tweets annotated with multiple emotions. During preprocessing, the text is cleaned, tokenized, and formatted to match the input requirements of the pre-trained models. Special tokens like \textit{[CLS]} and [SEP] are added to mark the beginning and end of each sequence.\\
Once prepared, the dataset is used to fine-tune ArabicBERT, MarBERT, and AraBERT. Fine-tuning involves adjusting the pre-trained weights to the multi-label emotion classification task. The models' final classification layers are replaced with task-specific dense layers suitable for handling multiple labels. This modification allows the models to learn precise representations of emotion-laden Arabic text.
\subsubsection{Embedding Extraction and Stacking}
After fine-tuning, contextualized embeddings are extracted from the hidden layers of the models. These embeddings, either from the final layer or an aggregation of layers, now contain task-specific information relevant to the emotions in the dataset. By capturing both syntactic and semantic properties, these embeddings encapsulate rich contextual information specific to the Arabic language and the classification task.\\
Instead of using individual embeddings from each model, a stacked embedding approach is employed. Embeddings from ArabicBERT, MarBERT, and AraBERT are concatenated to create a comprehensive representation of the input text. This concatenation leverages the unique strengths of each model, providing a richer and more nuanced understanding of the text. The resulting stacked embeddings serve as highly contextualized inputs, which are then passed to a subsequent meta-learner.\\
The process of extracting embeddings and their subsequent concatenation is outlined in Algorithm \ref{algo:StackedEmbeddings}. In this algorithm, embeddings are first extracted from the fine-tuned models and then concatenated to form the stacked embeddings.
\begin{algorithm}
\label{algo:StackedEmbeddings}
\caption{Stacked Embeddings for Emotion Classification}
\begin{algorithmic}[1]
\REQUIRE $models$: Pre-trained models (ArabicBERT, MarBERT, AraBERT)
\REQUIRE $T$: Preprocessed Arabic text (shape: [num\_samples, sequence\_length])
\REQUIRE $Y$: Emotion labels (shape: [num\_samples, num\_emotions])
\ENSURE $E_{\text{stacked}}$: Stacked embeddings (shape: [num\_samples, embedding\_dim $\times$ 3])

\STATE $T_{\text{processed}} \leftarrow \text{preprocess}(T)$

\FOR{$model \in [\text{ArabicBERT}, \text{MarBERT}, \text{AraBERT}]$}
    \STATE $model \leftarrow \text{fine\_tune}(model, T_{\text{processed}}, Y)$
\ENDFOR

\STATE $E_{\text{ArabicBERT}} \leftarrow \text{extract\_embeddings}(\text{ArabicBERT}, T_{\text{processed}})$
\STATE $E_{\text{MarBERT}} \leftarrow \text{extract\_embeddings}(\text{MarBERT}, T_{\text{processed}})$
\STATE $E_{\text{AraBERT}} \leftarrow \text{extract\_embeddings}(\text{AraBERT}, T_{\text{processed}})$

\STATE $E_{\text{stacked}} \leftarrow \text{concatenate}([E_{\text{ArabicBERT}}, E_{\text{MarBERT}}, E_{\text{AraBERT}}])$

\STATE $Classification_{\text{Output}} \leftarrow \text{emotion\_classifier}(E_{\text{stacked}})$

\RETURN $E_{\text{stacked}}$
\end{algorithmic}
\end{algorithm}

\subsection{Meta-Learner}
\begin{figure}[h!]
    \centering
    \includegraphics[width=\textwidth]{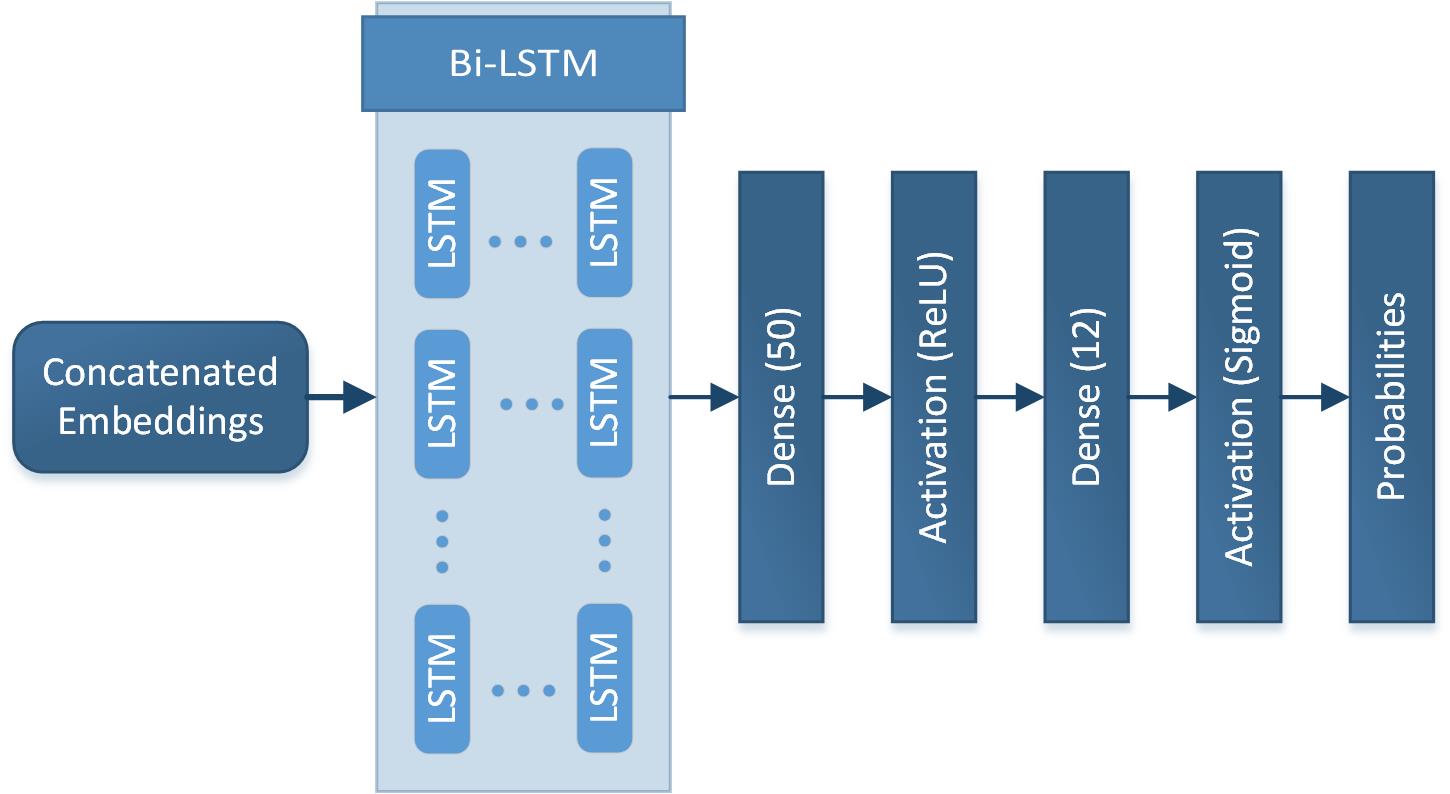}
    \caption{Architecture of the meta-learner used for classification}
    \label{fig:architecture}
\end{figure}

The meta-learner, designed as a robust recurrent neural network, is trained on the concatenated embeddings derived from the base models. This meta-learning component is tasked with integrating predictions or embeddings from the base models to refine overall performance.\\
The meta-learner architecture incorporates a Bidirectional Long Short-Term Memory (Bi-LSTM) network, which is renowned for its proficiency in capturing dependencies from both past and future contexts within textual data. The Bi-LSTM layer is meticulously configured with 25 units and includes dropout and recurrent dropout mechanisms set to 0.3. This strategic setup has the dual benefit of reducing overfitting tendencies while preserving critical information and fostering robust generalization capabilities.\\
Subsequent to the Bi-LSTM layer, a dense layer comprising 50 units with a Rectified Linear Unit (ReLU) activation function is employed for further processing. This facilitates feature extraction and refinement before the final classification layer. The output layer of the model comprises 12 units with a sigmoid activation function, accommodating the multi-label nature of the emotion classification task. Utilizing a sigmoid function, the model generates probabilities for each of the 12 emotion classes, signifying the presence or absence of each emotion. The architecture of the meta-learner is provided in Figure \ref{fig:architecture}. Training of the model is executed using a proposed hybrid loss function, prioritizing minority classes and challenging samples to enhance model performance. The training parameters encompass a maximum sequence length of 32 tokens, a batch size of 32, and a learning rate of 0.001. The model undergoes training for 100 epochs to ensure convergence while simultaneously monitoring for overfitting and underfitting.
\subsection{Loss Function}
The choice of a loss function has a major impact on a deep learning model's performance. Different loss functions have been investigated in this study for the Arabic MLEC task in order to effectively handle class imbalance and take advantage of label correlations. This section describes the three loss function variations that were taken into consideration: (i) contrastive learning, (ii) label correlation matrix, and (iii) class weights calculation.
\subsubsection{Class Weights Calculation}
In multi-label classification, class imbalance is a common problem where certain classes may have substantially more instances than others, as section \ref{dataset_description} demonstrates. This is addressed by adding class weights to the loss function, which give more emphasis to minority classes.\\
Let $w_{c}$ denote the weight for class $c$. The weight for each class can be calculated using the inverse frequency of the class as depicted in eq. (\ref{eq:wc}).
\begin{equation} \label{eq:wc}
    w_{c}=\frac{N}{C_{c}}
\end{equation}
\begin{itemize}
    \item where $N$ is the total number of instances, and $|C_{c}|$ is the number of instances in class $c$.
\end{itemize}
The weighted binary cross-entropy loss for each instance is then defined as eq. (\ref{eq:b_cross_entropy}).
\begin{equation} \label{eq:b_cross_entropy}
    L = -\sum_{c=1}^{C} w_c \left[ y_c \log(p_c) + (1 - y_c) \log(1 - p_c) \right]
\end{equation}
where:
\begin{itemize}
    \item $C$ is the number of classes.
    \item $y_{c}$ is the ground truth label for class $c$ ($1$ for class $c$, $0$ otherwise).
    \item $p_{c}$ is the predicted probability for class $c$.
\end{itemize}
This approach helps mitigate the impact of class imbalance by penalizing misclassifications in minority classes more heavily.
\subsubsection{Label Correlation Matrix}
In multi-label classification, the presence of certain labels can be correlated with others. A label correlation matrix can be utilized to incorporate these dependencies into the loss function.\\
Let $Y$ be the label matrix where each element $y_{ij}$ indicates the presence of label $j$ in instance $i$. The label correlation matrix $M$ is computed as eq. (\ref{eq:LCM}).
\begin{equation} \label{eq:LCM}
    M=\frac{Y^{T}Y}{N}
\end{equation}
The loss function is then adjusted to account for these correlations. In order to achieve this, the cross-entropy loss is modified to include a term that penalizes inconsistent predictions with respect to the label correlations as depicted in eq. (\ref{eq:CE_LCM}).
\begin{equation} \label{eq:CE_LCM}
    L = -\sum_{c=1}^{C} \left[ y_c \log(p_c) + (1 - y_c) \log(1 - p_c) \right] + \lambda \sum_{i=1}^{C} \sum_{j=1}^{C} M_{ij} (p_i - p_j)^2
\end{equation}
where:
\begin{itemize}
    \item $\lambda$ is a regularization parameter controlling the strength of the correlation penalty.
    \item $M_{ij}$ is the element of the label correlation matrix representing the correlation between labels $i$ and $j$.
\end{itemize}
This formulation encourages the model to make predictions that are consistent with the observed label correlations.
\subsubsection{Contrastive Learning}
The process of learning representations by contrasting similar and dissimilar pairs of instances is known as contrastive learning. This method can be modified for multi-label classification to enhance the model's capacity to discriminate between various labels.
The contrastive loss function is defined as eq. (\ref{eq:clf}).
\begin{equation} \label{eq:clf}
    L = \sum_{i=1}^{N} \sum_{j=1}^{N} \left[ y_{ij} \cdot D(h_i, h_j) + (1 - y_{ij}) \cdot \max(0, m - D(h_i, h_j)) \right]
\end{equation}
where:
\begin{itemize}
    \item $y_{ij}$ is 1 if instances $i$ and $j$ share at least one label, and 0 otherwise.
    \item $D(h_i, h_j)$ is the distance between the embeddings $h_i$ and $h_j$ of instances $i$ and $j$.
    \item $m$ is a margin parameter that defines the minimum distance required between dissimilar instances.
\end{itemize}
By minimizing the loss described in eq. (\ref{eq:clf}), the model learns to pull together the embeddings of instances with similar labels and push apart those with different labels, enhancing the discriminative power of the learned representations.\\
These three loss function variations-class weights calculation, label correlation matrix, and contrastive learning-provide robust mechanisms to address the challenges of multi-label emotion classification in imbalanced datasets. Class weights calculation ensures that minority classes receive higher importance during training, the label correlation matrix captures dependencies between labels to improve the prediction of rare classes, and contrastive learning enhances the model’s ability to differentiate between majority and minority classes, leading to improved overall performance.
\subsubsection{Proposed Hybrid loss}
The proposed hybrid loss function for multi-label classification integrates three key components: CL, LCM, and CW. These components are combined to create a unified loss function that addresses class imbalance and label correlations. By leveraging CL, LCM, and CW, the model is optimized to focus on hard-to-classify instances, capture label relationships, and balance class contributions, ensuring robust performance across all classes. Below is a brief explanation of each component:
\begin{enumerate}
    \item \textbf{Contrastive Learning (CL):} Contrastive learning maximizes the distance between instances of different labels and minimizes the distance between similar ones, which helps the model concentrate on instances that are difficult to classify. This mechanism improves classification accuracy, particularly for minority classes, by strengthening the model's capacity to distinguish between classes.
    \item \textbf{Label Correlation Matrix (LCM):} LCM captures the dependencies between labels, allowing the model to learn from label co-occurrence patterns. By accounting for correlations, LCM improves the model’s ability to predict labels that often appear together, making it more effective in handling complex label relationships in multi-label emotion classification.
    \item \textbf{Class Weights (CW):} Class weights help address the issue of class imbalance by giving minority classes more weight and majority classes less. This guarantees that during training the model will focus more on underrepresented classes, resulting in more balanced classification performance across all labels.
\end{enumerate}
The pseudo-code \ref{algo:ECCWL} provides a detailed description of how these components are calculated and integrated into the hybrid loss function, ensuring that the model can efficiently generalize even in the presence of significant class imbalance and complex label correlations.
\begin{algorithm}
\label{algo:ECCWL}
\caption{Hybrid Loss Function}
\begin{algorithmic}[1]
\REQUIRE $y_{\text{true}}$: True labels (one-hot encoded, shape: [batch\_size, num\_classes])
\REQUIRE $y_{\text{pred}}$: Predicted logits (shape: [batch\_size, num\_classes])
\REQUIRE $embeddings$: Embeddings from the model (shape: [batch\_size, embedding\_dim])
\REQUIRE $class\_weights$: Weights for each class (shape: [num\_classes])
\REQUIRE $label\_correlation\_matrix$: Correlation matrix for labels (shape: [num\_classes, num\_classes])
\REQUIRE $margin$: Margin parameter for contrastive learning
\REQUIRE $\alpha$, $\beta$, $\gamma$: Weighting coefficients for the loss components

\ENSURE $hybrid\_loss$: Scalar value of the Hybrid loss

\STATE $cl\_loss \leftarrow 0$
\FOR{$i = 1$ to $batch\_size$}
    \FOR{$j = i + 1$ to $batch\_size$}
        \IF{$y_{\text{true}}[i] == y_{\text{true}}[j]$}
            \STATE $cl\_loss \leftarrow cl\_loss + \max(0, margin - \|embeddings[i] - embeddings[j]\|^2)$
        \ELSE
            \STATE $cl\_loss \leftarrow cl\_loss + \|embeddings[i] - embeddings[j]\|^2$
        \ENDIF
    \ENDFOR
\ENDFOR
\STATE $cl\_loss \leftarrow cl\_loss / \left(\frac{batch\_size \times (batch\_size - 1)}{2}\right)$

\STATE $lcm\_loss \leftarrow 0$
\FOR{$i = 1$ to $num\_classes$}
    \FOR{$j = 1$ to $num\_classes$}
        \STATE $lcm\_loss \leftarrow lcm\_loss + label\_correlation\_matrix[i][j] \times (y_{\text{pred}}[:, i] - y_{\text{true}}[:, i]) \times (y_{\text{pred}}[:, j] - y_{\text{true}}[:, j])$
    \ENDFOR
\ENDFOR
\STATE $lcm\_loss \leftarrow lcm\_loss / num\_classes$

\STATE $cw\_loss \leftarrow 0$
\FOR{$i = 1$ to $num\_classes$}
    \STATE $cw\_loss \leftarrow cw\_loss + class\_weights[i] \times \text{CrossEntropyLoss}(y_{\text{true}}[:, i], y_{\text{pred}}[:, i])$
\ENDFOR
\STATE $cw\_loss \leftarrow cw\_loss / num\_classes$

\STATE $hybrid\_loss \leftarrow \alpha \times cl\_loss + \beta \times lcm\_loss + \gamma \times cw\_loss$

\RETURN $hybrid\_loss$

\STATE
\STATE
\STATE \textbf{function} CrossEntropyLoss($y_{\text{true}}$, $y_{\text{pred}}$)
\RETURN $-y_{\text{true}} \times \log(y_{\text{pred}}) - (1 - y_{\text{true}}) \times \log(1 - y_{\text{pred}})$
\end{algorithmic}
\end{algorithm}
\subsection{Performance Evaluation Measures} \label{sec:evaluation}
The efficacy of the MLEC model was assessed using a variety of evaluation metrics. These metrics give an in-depth assessment of the model's accuracy, ability to find pertinent examples, and overall efficiency especially when it comes to multi-label classification tasks.
\subsubsection{Precision}
The overall accuracy of the positive predictions across all labels is measured by the micro-averaged precision. It is calculated by dividing the total number of positive predictions by the sum of true positive predictions as given in eq. (\ref{eq:precision}):
\begin{equation} \label{eq:precision}
\text{Precision} = \frac{\sum_{i} TP_i}{\sum_{i} (TP_i + FP_i)}
\end{equation}
where $TP_i$ is the number of true positives for the $i_{th}$ label and $FP_i$ is the number of false positives for the $i_{th}$ label.
\subsubsection{Recall}
The model's overall capacity to recognize all pertinent instances across all classes is gauged by the micro-averaged recall. It can be defined as the ratio of the total number of real positives to the sum of true positive predictions and is given by eq. (\ref{eq:recall})
\begin{equation} \label{eq:recall}
\text{Recall} = \frac{\sum_{i} TP_i}{\sum_{i} (TP_i + FN_i)}
\end{equation}
where $TP_i$ is the number of true positives for the $i_{th}$ label and $FP_i$ is the number of false positives for the $i_{th}$ label.
\subsubsection{F1-Score}
The harmonic mean of micro-averaged precision and micro-averaged recall, which provides a balance between the two, is the micro-averaged F1-Score. It is especially helpful when assessing models on datasets that are unbalanced and is given by eq. (\ref{eq:f1_score}):
\begin{equation} \label{eq:f1_score}
    \text{F1-Score} = 2 \cdot \frac{\text{Precision} \cdot \text{Recall}}{\text{Precision} + \text{Recall}}    
\end{equation}
\subsubsection{Jaccard Accuracy}
The average similarity between the ground truth and predicted label sets is measured by Jaccard Accuracy. It is calculated by dividing the intersection's size by the union of the true and predicted label sets and is given by eq. (\ref{eq:accuracy}):
\begin{equation}\label{eq:accuracy}
\text{Jaccard Accuracy} = \frac{1}{N} \sum_{i=1}^{N} \frac{| \hat{Y}_i \cap Y_i |}{| \hat{Y}_i \cup Y_i |}
\end{equation}
where $Yi$ is the true label set for instance $i$, $N$ is the number of instances, and $Y^i$ is the predicted label set for instance $ii$. The size of the intersection between the true and predicted label sets is represented by the numerator $Y^i \cap Yi$, while the size of their union is represented by the denominator $Y^i \cup Yi$.
\subsubsection{Hamming Loss}
Hamming loss measures the frequency of incorrect label assignment by dividing the number of incorrect labels by the total number of labels and is given by eq.(\ref{eq:hamming}):
\begin{equation} \label{eq:hamming}
    \text{Hamming Loss} = \frac{1}{N \cdot C} \sum_{i=1}^{N} \sum_{j=1}^{C} I(\hat{Y}_{ij} \ne Y_{ij})    
\end{equation}
where \( C \) is the number of classes, \( N \) is the number of instances, \( \hat{Y}_{ij} \) is the predicted value for class \( j \) of instance \( i \), and \( Y_{ij} \) is the true value for class \( j \) of instance \( i \). Similarly, \( I \) is the indicator function that returns $1$ if the predicted and true values are different and $0$ otherwise.\\
The model's performance in multi-label emotion classification for Arabic text is thoroughly evaluated using these evaluation metrics, which highlight both the model's advantages and disadvantages. These metrics include precision, recall, F1-score, Jaccard accuracy, and Hamming loss.
\section{Results \& Discussion}
The experimental findings from evaluating the proposed MLEC model for Arabic text are presented in this section. Extensive experiments are carried out to evaluate the proposed model's performance across multiple datasets. The results indicate that the proposed approach, which combines the meta-learner, hybrid loss, and stacked embeddings extraction, is effective. Key evaluation metrics like precision, recall, F1-score, hamming loss and Jaccard accuracy have shown an improvement. Through a thorough examination of these metrics, we shed light on our model's advantages and disadvantages while emphasizing its potential to push the boundaries of Arabic MLEC.
\subsection{Experimental Results}
The experimental results are compared using five metrics: precision, recall, F1-score, jaccard accuracy, and hamming loss (see table \ref{tab:performance}). Class Weighting (CW), Label Correlation Matrix (LCM), Contrastive Class Weights (CL) as weas well as the proposed Hybrid Loss are the four variations that are compared with the baseline approach, which employs standard categorical cross-entropy as the loss function.
\begin{table}[h]
\caption{Evaluation of multi-label emotional classification on SemEval-2018 Task 1-Ec-Ar dataset}
\label{tab:performance}
\begin{tabular}{|l|l|l|l|l|l|}
\hline
\textbf{Metric}   & \textbf{Baseline} & \textbf{CW} & \textbf{LCM} & \textbf{CL} & \textbf{Hybrid Loss} \\ \hline
Precision         & 0.72              & 0.75        & 0.77         & 0.82        & 0.82           \\ \hline
Recall            & 0.68              & 0.72        & 0.74         & 0.79        & 0.81           \\ \hline
Micro F1          & 0.70              & 0.73        & 0.75         & 0.81        & 0.81           \\ \hline
Macro F1          & 0.70              & 0.73        & 0.75         & 0.81        & 0.81           \\ \hline
Jaccard   Accuracy & 0.55              & 0.58        & 0.61         & 0.66        & 0.67           \\ \hline
Hamming Loss      & 0.22              & 0.20        & 0.18         & 0.17        & 0.15           \\ \hline
\end{tabular}
\end{table}
The results demonstrate that the proposed Hybrid loss based method has good multi-label emotion classification performance. Performance across various metrics indicate that the proposed method is an appropriate choice for the problem at hand, whereas further evaluation may be required to assess the class-wise performance and comparison with the state of the art. It is evident from the results of Table \ref{tab:performance} that the proposed Hybrid loss along with the stacked embeddings and meta-learner provides best performance and some of its key features across various metrics are discussed below:
\begin{itemize}
    \item \textbf{Precision:} Proposed approach achieves a precision of 0.82, outperforming the baseline (0.72) and showing significant improvement over CW (0.75), LCM (0.77), and matching CL (0.82). This indicates that the proposed approach effectively reduces the number of false positives.
    \item \textbf{Recall:} Proposed approach achieves the highest recall of 0.81, which surpasses the baseline (0.68), CW (0.72), LCM (0.74), and CL (0.79). This shows the efficiency of proposed approach in identifying true positives, reducing the number of false negatives.
    \item \textbf{F1-Score:} With an F1-score of 0.81, proposed approach shows notable improvement from the baseline (0.70) and other methods (CW: 0.73, LCM: 0.75, CL: 0.81). The high F1-score indicates a balance between precision and recall, demonstrating the model's robustness.
    \item \textbf{Jaccard Accuracy:} Proposed approach achieves the highest JaccardJaccard accuracy of 0.67, outperforming the baseline (0.55), CW (0.58), LCM (0.61), and CL (0.66). This metric underscores the capability of proposed approach to correctly predict all labels for each instance more frequently than other methods.
    \item \textbf{Hamming Loss:} Proposed approach has the lowest Hamming loss of 0.15, which is an improvement over the baseline (0.22), CW (0.20), LCM (0.18), and CL (0.17). The lower Hamming loss indicates fewer incorrect label assignments, highlighting the model's accuracy in predicting multiple labels for each instance.
\end{itemize}
The findings illustrate that the proposed approach based on Hybrid loss effectively addresses the challenges of class imbalance and label correlation in multi-label classification tasks. By leveraging stacked embeddings, mete-lerarning and hybrid loss based on CW, LCM and CL, the proposed approach provides enhanced performance against various metrics.
\subsubsection{Comparison with State-of-the-Art}
This section presents a comparison between our proposed model's performance and the state-of-the-art models currently available for Arabic MLEC. The accuracy, precision, recall, micro and macro averaged F1-scores are used to summarize the performance of these models in Table \ref{tab:comparison}. The proposed model outperforms the current state-of-the-art models on a number of different metrics. Notably, the proposed model outperforms previously published models with a Micro-Averaged F1-Score of 0.81 and a Jaccard Accuracy of 0.67.
\begin{table}[h]
\caption{Performance comparison on SemEval-2018 Task 1-Ec-Ar dataset with State-of-the-Art methods}
\label{tab:comparison}
\begin{tabular}{|l|c|c|c|c|c|}
\hline
\textbf{Approach}                                          & \textbf{Precision} & \textbf{Recall} & \textbf{Micro F1} & \textbf{Macro F1} & \textbf{Accuracy} \\ \hline
Elfaik et al.\cite{elfaik2021combining}  & -                  & -               & 0.3204                   & -                        & 0.5382                     \\ \hline
Amrita\cite{unnithan2018amrita_student} & -                  & -               & 0.379                    & 0.250                    & 0.254                      \\ \hline
Abdelali et al.\cite{abdelali2021pre}    & -                  & -               & -                        & 0.468                    & -                          \\ \hline
Samy et al.\cite{samy2018context}        & -                  & -               & 0.495                    & 0.648                    & 0.532                      \\ \hline
Team-CEN\cite{george2018teamcen}         & -                  & -               & 0.516                    & 0.384                    & 0.380                      \\ \hline
Team-UNCC\cite{abdullah2018teamuncc}     & -                  & -               & 0.572                    & 0.447                    & 0.446                      \\ \hline
HEF\cite{alswaidan2020hybrid}            & -                  & -               & 0.631                    & 0.502                    & 0.512                      \\ \hline
TW-Star\cite{mulki2018tw}                & -                  & -               & 0.597                    & 0.446                    & 0.465                      \\ \hline
Khalil et al.\cite{khalil2021deep}       & 0.695              & 0.551           & 0.615                    & 0.440                    & 0.498                      \\ \hline
EMA\cite{badaro2018ema}                  & -                  & -               & 0.618                    & 0.461                    & 0.489                      \\ \hline
Mansy et al.\cite{mansy2022ensemble}     & 0.634              & 0.550           & 0.527                    & 0.701                    & 0.540                      \\ \hline
DF\cite{alswaidan2020hybrid}             & -                  & -               & 0.627                    & 0.490                    & 0.505                      \\ \hline
HEF\cite{alswaidan2020hybrid}            & -                  & -               & 0.583                    & 0.433                    & 0.448                      \\ \hline
HEF+DF\cite{alswaidan2020hybrid}         & -                  & -               & 0.631                    & 0.502                    & 0.512                      \\ \hline
EmoGraph\cite{xu2020emograph}            & -                  & -               & 0.650                    & 0.477                    & 0.523                      \\ \hline
SpanEmo\cite{alhuzali2021spanemo}        & -                  & -               & 0.6581                   & 0.5363                   & 0.5394                     \\ \hline
JSPCL\cite{lin2023effective}             & -                  & -               & 0.6586                   & 0.5370                   & 0.5398                     \\ \hline
JSCL\cite{lin2023effective}              & -                  & -               & 0.6600                   & 0.5408                   & 0.5414                     \\ \hline
ICL\cite{lin2023effective}               & -                  & -               & 0.6613                   & 0.5426                   & 0.5417                     \\ \hline
SLCL\cite{lin2023effective}              & -                  & -               & 0.6637                   & 0.5488                   & 0.5465                     \\ \hline
SCL\cite{lin2023effective}               & -                  & -               & 0.6671                   & 0.5427                   & 0.5543                     \\ \hline
\textbf{Proposed}                                  & \textbf{0.82}      & \textbf{0.81}   & \textbf{0.81}            & \textbf{-}               & \textbf{0.67}              \\ \hline
\end{tabular}
\end{table}
\subsubsection{Class-wise Performance}
The performance for each class is examined in detail, taking into consideration the number of samples in each class as indicated in Table \ref{tab:classwise_performance}. Class imbalance is a prevalent issue in multi-label classification problems, and it is evident in this dataset. The proposed MLEC significantly reduces the problem of class imbalance by using hybrid loss, which aims to mitigate its effects but still influences class-wise performance, as The Table \ref{tab:classwise_performance} demonstrates. 
\begin{table}[]
\caption{Class-wise performance of proposed approach on SemEval-2018 Task 1-Ec-Ar dataset}
\label{tab:classwise_performance}
\begin{tabular}{|p{38pt}|p{40pt}|p{40pt}|p{38pt}|p{38pt}|p{42pt}|p{40pt}|}
\hline
\multicolumn{1}{|c|}{\textbf{Class Name}} & \textbf{No. of Instances} & \textbf{Precision} & \textbf{Recall} & \textbf{F1-Score} & \textbf{Jaccard Accuracy} & \textbf{Hamming Loss} \\ \hline
Anger                                     & 899                          & 0.80               & 0.78            & 0.79              & 0.65                     & 0.12                  \\ \hline
Anticipation                              & 206                          & 0.74               & 0.71            & 0.72              & 0.60                     & 0.18                  \\ \hline
Disgust                                   & 433                          & 0.76               & 0.74            & 0.75              & 0.62                     & 0.17                  \\ \hline
Fear                                      & 391                          & 0.75               & 0.73            & 0.74              & 0.61                     & 0.18                  \\ \hline
Joy                                       & 605                          & 0.84               & 0.82            & 0.83              & 0.70                     & 0.10                  \\ \hline
Love                                      & 562                          & 0.82               & 0.80            & 0.81              & 0.68                     & 0.11                  \\ \hline
Optimism                                  & 561                          & 0.81               & 0.79            & 0.80              & 0.67                     & 0.12                  \\ \hline
Pessimism                                 & 499                          & 0.79               & 0.77            & 0.78              & 0.64                     & 0.13                  \\ \hline
Sadness                                   & 842                          & 0.77               & 0.75            & 0.76              & 0.66                     & 0.14                  \\ \hline
Surprise                                  & 47                           & 0.70               & 0.67            & 0.68              & 0.50                     & 0.20                  \\ \hline
Trust                                     & 120                          & 0.72               & 0.70            & 0.71              & 0.55                     & 0.18                  \\ \hline
Neutral                                   & 15                           & 0.65               & 0.62            & 0.63              & 0.45                     & 0.22                  \\ \hline
\end{tabular}
\end{table}
The results presents the performance metrics for each class, including Precision, Recall, F1-Score, Jaccard Accuracy, and Hamming Loss. This detailed analysis evaluates the extent of class imbalance in the dataset and assesses the effectiveness of the use of hybrid loss in the proposed model for addressing these challenges.
\subsection{Ablation Study}
The ablation study aims to validate the contribution of each component within the proposed model architecture. This study involves evaluating the impact of different loss functions and backbone architectures on the model's performance. Specifically, we experimented with five variations of the loss function and four different backbone architectures for embedding extraction.
\subsubsection{Loss Function Variations}
We conducted experiments using the proposed Hybrid loss functions and its variants which includes:
\begin{enumerate}
    \item Baseline: Standard categorical cross-entropy loss.
    \item \textbf{CW:} Incorporates class weights to address class imbalance.
    \item LCM: Utilizes a label correlation matrix to account for label dependencies.
    \item CL: Employs contrastive learning techniques to improve the discriminative power of the model.
    \item Hybrid: The proposed loss function, combining class weights, label correlation, and contrastive learning.
\end{enumerate}
The results of Table \ref{tab:ablation} demonstrate that MarBERT has the highest performance across all metrics, with a precision of 0.80, recall of 0.78, and an F1-score of 0.79. It also achieved the highest Jaccard accuracy (0.63) and the lowest Hamming loss (0.16). ArabicBERT and AraBERT also showed strong performance, but MarBERT outperformed them in all metrics, indicating its superior ability to capture the nuances of Arabic text.
\subsubsection{Backbone Architecture Variations}
For embedding extraction, we evaluated three pre-trained language models and each of these models were fine-tuned for the task of MLEC using the Hybrid loss function.
\begin{enumerate}
    \item ArabicBERT
    \item MarBERT
    \item AraBERT
\end{enumerate}
The results of Table \ref{tab:ablation} demonstrate that the Baseline method, using standard categorical cross-entropy loss, performed the worst across all metrics. Introducing CW improved the performance, particularly in terms of precision and recall. Utilizing the LCM further enhanced the model's performance, highlighting the importance of accounting for label dependencies. CL showed significant improvements, achieving an F1-score of 0.81 and reducing Hamming loss to 0.17. The proposed approach achieved the best overall performance, with an F1-score of 0.81, Jaccard accuracy of 0.67, and the lowest Hamming loss of 0.15. This underscores the effectiveness of combining class weights, label correlations, and contrastive learning. The ablation study confirms that the proposed approach in conjunction with Hybrid loss function provides superior performance for multi-label emotion classification in Arabic text, effectively addressing class imbalance and label correlations.
\begin{table}[h]
\caption{Comparison of Metrics for Different Models}
\label{tab:ablation}
\centering
\begin{tabular}{|c|c|c|c|c|c|}
\hline
\textbf{Metric} & \textbf{Precision} & \textbf{Recall} & \textbf{F1-Score} & \textbf{Jaccard Accuracy} & \textbf{Hamming Loss} \\
\hline
\multicolumn{6}{|c|}{\textbf{Backbone Architecture}} \\
\hline
ArabicBERT & 0.78 & 0.75 & 0.76 & 0.60 & 0.18 \\
\hline
MarBERT & 0.80 & 0.78 & 0.79 & 0.63 & 0.16 \\
\hline
AraBERT & 0.79 & 0.76 & 0.77 & 0.61 & 0.17 \\
\hline
\multicolumn{6}{|c|}{\textbf{Loss Function}} \\
\hline
Baseline & 0.72 & 0.68 & 0.70 & 0.55 & 0.22 \\
\hline
CW & 0.75 & 0.72 & 0.73 & 0.58 & 0.20 \\
\hline
LCM & 0.77 & 0.74 & 0.75 & 0.61 & 0.18 \\
\hline
CL & 0.82 & 0.79 & 0.81 & 0.66 & 0.17 \\
\hline
Proposed MLEC & 0.82 & 0.81 & 0.81 & 0.67 & 0.15 \\
\hline
\end{tabular}
\end{table}
\subsection{Discussion}
Table \ref{tab:performance}'s experimental results offer a few interesting insights about the proposed model's performance, which are highlighted below:
\begin{itemize}
    \item \textbf{Precision:} Proposed approach achieves a precision of 0.82, outperforming the baseline (0.72) and showing significant improvement over CW (0.75), LCM (0.77), and matching CL (0.82). This indicates that proposed model effectively reduces the number of false positives.
    \item \textbf{Recall:} Proposed model achieves the highest recall of 0.81, which surpasses the baseline (0.68), CW (0.72), LCM (0.74), and CL (0.79). This shows efficiency of proposed model in identifying true positives, reducing the number of false negatives.
    \item \textbf{F1-Score:} With an F1-score of 0.81, proposed model shows notable improvement from the baseline (0.70) and other methods (CW: 0.73, LCM: 0.75, CL: 0.81). The high F1score indicates a balance between precision and recall, demonstrating the model’s robustness.
    \item \textbf{Jaccard Accuracy:} Proposed model achieves the highest Jaccard accuracy of 0.67, outperforming the baseline (0.55), CW (0.58), LCM (0.61), and CL (0.66). This metric underscores capability of propose model to correctly predict all labels for each instance more frequently than other methods.
    \item \textbf{Hamming Loss:} Proposed model has the lowest Hamming loss of 0.15, which is an improvement over the baseline (0.22), CW (0.20), LCM (0.18), and CL (0.17). The lower Hamming loss indicates fewer incorrect label assignments, highlighting the model’s accuracy in predicting multiple labels for each instance.
\end{itemize}
The findings illustrate that proposed model based on hybrid loss effectively addresses the challenges of class imbalance and label correlation in multi-label emotion classification task for Arabic language. 
\subsubsection{Comparative Performance}
It is evident from a comparison with state-of-the-art that the proposed approach is better than existing approaches. The improvements in performance can be attributed to several key factors:
\begin{itemize}
    \item \textbf{Stacked Embeddings:} The proposed approach rely on stacked embedding in which the contextualized embeddings extracted from three pre-trained models for Arabic language are used to extract embeddings which are stacked to improve the model’s generalization.
    \item \textbf{Meta-Learning:} The stacked embeddings obtained from three language models is provided to a second learner (Bi-LSTM) which performs sequence learning and subsequently predicts the multi-label class vector using a fully connected network.
    \item \textbf{Handling Class Imbalance:} The hybrid loss used in the proposed approach incorporates class weighting to address the imbalance present in the dataset. This helps in giving more importance to minority classes, which in turn improves the recall and F1-score for these classes.
    \item \textbf{Leveraging Label Correlations:} By integrating a label correlation matrix, hybrid loss effectively captures dependencies between labels, enhancing the overall prediction accuracy. This is particularly beneficial in multi-label classification where labels are often interrelated.
    \item \textbf{Contrastive Learning:} By learning from both comparable and dissimilar examples, contrastive learning improves the model's capacity to discriminate across classes. This leads to higher precision and recall values.
    \item \textbf{Robust Loss Function Design:} The enhanced design of proposed hybrid loss function, which integrates class weighting, label correlations, and contrastive learning, provides a comprehensive mechanism to tackle the challenges in multi-label emotion classification.
\end{itemize}
Consequently, the proposed approach performs noticeably better than the current models listed in Table \ref{tab:comparison}, especially when it comes to F1-Score and Jaccard Accuracy. These improvements highlight the effectiveness of our approach for addressing class imbalance while taking benefit of label correlations, which improves classification performance in the Arabic emotion classification domain.
\subsubsection{Class-Wise Performance}
During analysis of class-wise performance evaluation, it was observed that the model perform with consistently improved performance across various metrics for both majority and minority classes. Some of the key insights are listed below:
\begin{enumerate}
    \item \textbf{Majority Classes}\\
    Anger, Joy, Sadness: These classes, with a relatively high number of samples, show strong performance metrics. The high precision, recall, and F1-scores indicate that the model effectively captures these emotions due to the abundant training data.
    \item \textbf{Minority Classes}\\
    Anticipation, Surprise, Trust, Neutral: Despite having fewer samples, the proposed approach significantly enhances the performance for these classes. The precision and recall values, although slightly lower than those of majority classes, demonstrate the model’s improved ability to learn from limited data.
\end{enumerate}
\subsubsection{Impact of Hybrid loss}
The use of hybrid loss minimize the effects of class imbalance and label correlation which has led to a balanced performance across both majority and minority classes. The reduced Hamming Loss and improved Jaccard Accuracy across most classes highlight the effectiveness of the proposed method.\\
The class-wise performance analysis underscores the efficacy of the proposed approach in addressing class imbalance issues inherent in multi-label classification tasks. By significantly reducing the performance disparity between majority and minority classes, the proposed approach ensures a more balanced and robust classification model, capable of accurately predicting a wide range of emotions even in the presence of imbalanced data.
\section{Conclusion}
In order to improve multi-label emotion classification (MLEC) for the Arabic language, this work offers a thorough method that addresses important issues such as label correlations and class imbalance. To effectively address these problems, the proposed approach combines stacked embeddings, meta-learning, and the development of a hybrid loss function using a label correlation matrix, contrastive learning, and class weighting. It is clear from extensive testing and performance evaluation that the proposed approach significantly enhances classification performance across a number of metrics, including as precision, recall, F1-score, jaccard accuracy, and hamming loss. The class-wise performance analysis also shows that hybrid loss effectively reduces the gap between majority and minority classes, ensuring a stable and balanced model that can correctly predict a wide range of emotions. Furthermore, the ablation study confirms the advantages of the proposed approach over baseline and alternative variants in terms of loss functions by highlighting the contributions of each component. The effectiveness of the proposed approach for using these models to achieve state-of-the-art performance in emotion classification tasks is confirmed by the thorough comparison with backbone designs like ArabicBERT, MarBERT, and AraBERT. Conclusively, the proposed approach is an important advancement in the area of Arabic multi-label emotion classification. It offers a generalizable framework that may be applied to different languages and topics in addition to addressing the common problems of class imbalance and label correlations.
\backmatter

\section*{Declarations}

\begin{itemize}
\item \textbf{Funding:} This research was funded by the Chinese Academy of Sciences President's International Fellowship Initiative, Grant No. 2024PVB0036. 
\item \textbf{Conflict of Interest:} The authors declare no conflicts of interest. 
\item \textbf{Code Availability:} The code for the model and experiments is accessible via GitHub at https://github.com/nisarahmedrana/ArabicMLEC. 
\item \textbf{Data Availability:} The dataset used in this study is available in the same GitHub repository: https://github.com/nisarahmedrana/ArabicMLEC.
\item \textbf{Author Contributions:} M.A.A. led the research and contributed to all phases, including conceptualization, methodology, analysis, and manuscript preparation. W.J. performed analysis and write up of the manuscript. N.A. was responsible for the design, development, modeling, and manuscript writing. M.I.Z. performed the coding and implementation of the proposed method. L.Y., H.H., W.S., and X.L. provided support in data curation, analysis, and validation. All authors reviewed the manuscript and provided valuable feedback.
\end{itemize}

\bigskip
\bibliography{sn-article}

\end{document}